\documentclass[times]{elsarticle}

\usepackage{amssymb}
\usepackage{hyperref}
\usepackage{amsmath}
\usepackage{commath}
\usepackage{multirow}
\usepackage{makecell}
\usepackage{csquotes}

\journal{Pattern Recognition}

\begin{document}

\begin{frontmatter}




\title{Ensemble Learning for Fusion of Multiview Vision with Occlusion and Missing Information: \\ Framework and Evaluations with Real-World Data and Applications in Driver Hand Activity Recognition}


\author[1]{Ross Greer\corref{cor1}} 

\cortext[cor1]{Corresponding author:}
\ead{regreer@ucsd.edu}

\author[1]{Mohan M. Trivedi}

\affiliation[1]{organization={University of California San Diego},
                addressline={Laboratory for Intelligent \& Safe Automobiles, 9500 Gilman Drive}, 
                city={La Jolla}, 
                state={California},
                country={USA}}

\begin{abstract}
Multi-sensor frameworks provide opportunities for ensemble learning and sensor fusion to make use of redundancy and supplemental information, helpful in real-world safety applications such as continuous driver state monitoring which necessitate predictions even in cases where information may be intermittently missing. We define this problem of intermittent instances of missing information (by occlusion, noise, or sensor failure) and design a learning framework around these data gaps, proposing and analyzing an imputation scheme to handle missing information. We apply these ideas to tasks in camera-based hand activity classification for robust safety during autonomous driving. We show that a late-fusion approach between parallel convolutional neural networks can outperform even the best-placed single camera model in estimating the hands' held objects and positions when validated on within-group subjects, and that our multi-camera framework performs best on average in cross-group validation, and that the fusion approach outperforms ensemble weighted majority and model combination schemes. 
\end{abstract}




\end{frontmatter}


\section{Introduction}

Manual (hand-related) activity is a significant source of crash risk while driving; driver distraction contributes to around 65\% of safety-critical events (crashes and near crashes) \cite{zangi2022driver}, and more than 3,000 deaths in 2022 \cite{cdc_driving}.  

Furthermore, given recent consumer adoption of early-stage autonomy in vehicles, driver hand activity has been shown to lead to various incidents even in these semi-autonomous vehicles. Drivers in vehicles supported by partial autonomy show high propensity to engage in distracting activities when supported by automation \cite{dingus2016driver} and show increased likelihood of crashes or near-crashes when engaged in distracting activity \cite{zangi2022driver}. Moreover, it is important to consider the manner of transitions when the driver must take manual control of semi-autonomous vehicles, as drivers demonstrate a slowness or inability to handle these control transitions safely when occupied with non driving-related tasks, often involving the hands \cite{naujoks2018partial} \cite{rangesh2021autonomous}.

Accordingly, analysis of hand position and hand activity occupation is a useful component to understanding a driver’s readiness to take control of a vehicle. Visual sensing through cameras provides a passive means of observing the hands, but its effectiveness varies depending on the camera location.

In this paper, we present a multi-camera sensing framework and machine learning solution, which we apply to the problem of robust driver state monitoring for autonomous driving safety. Our real-world, constrained application represents just one use case for this framework, as it can readily be extended to an ensemble of $N$ domain-agnostic data sources and models for similar tasks; accordingly, we provide both a domain-specific and a generalized formulation of the sensing framework and learning problem in the following sections. 

Consider an intelligent vehicle which classifies a driver's hand activity for a downstream safety application. By constraints imposed by vehicle manufacturing, we may have multiple cameras (in our case, four) which observe the driver from varying angles: head-on from the steering wheel, diagonally from the rearview mirror, diagonally from the dashboard, and peripherally from the central console. It is readily apparent that, depending on the driver's current position, there are instances where:
\begin{enumerate}
    \item Only one of the four cameras has any view of the driver's hands, or
    \item Multiple cameras have a view of the driver's hands.
\end{enumerate}

\begin{figure}
    \centering
    \includegraphics[width=\textwidth]{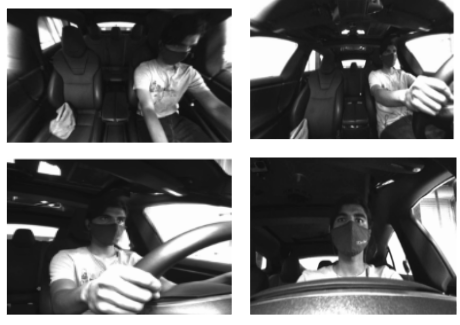}
    \caption{Multi-view images (clockwise) - rearview, dashboard center, steering view, and dashboard driver, which explain how hands can be missing in certain frames, causing an \textit{irregular redundancy}.} 
    \label{fig:capture}
\end{figure}


An ideal intelligent system would recognize which of the cases is present, and in the former, choose to use the visible information to make an estimate, and in the latter, form an estimate made with the joint information of the multiple views which may be helpfully redundant (both cameras observing the fingers grasping the wheel, from multiple directions) or supplemental (one camera observes the fingers clasped to the wheel rim, the other observes the wrist resting on the wheel center) to the task at hand. This redundancy is closely related to the concept of \textit{homogeneity} in \cite{liang2022foundations}. Because this redundant or supplemental information can be present or absent between instances, we refer to this particular ``missing data" phenomenon as \textit{irregular redundancy}. Conceptually, this is similar to situations where data streams which operate under noisy conditions or at different sample rates are provided as input to a model which must provide output despite lost frames due to sampling rate or corruption. 

More generally, we may describe this as a problem of \textit{sensor fusion}, by which we must handle data to best leverage the accompanying noise, variance, and redundancy between samples to create an optimal estimate. In this theme, here we pose our framework as a system in which we have multiple data sources of the same event, and our goal is to learn an optimal model which accurately estimates a property of the event. Here, we are left with a few choices:

\begin{enumerate}
    \item A model learned from one of these sources may tend to provide the best estimate, and we use only this source for future inference.
    \item Models learned independently from each of these sources can each provide an independent estimate, and we can interpret their respective estimates to reach a group-informed consensus estimate.
    \item A single model can be learned simultaneously between the sources, exploiting moments of redundancy and uncertainty in the data sources, such that the model provides an estimate with intelligence in selecting relevant features from data sources at any given instance dependent on the state of the other sources. 
\end{enumerate}

This question, described as the \textit{multimodal reasoning} problem \cite{liang2022foundations}, is thoroughly investigated in the work of Seeland and Mäder \cite{seeland2021multi}, as will be discussed on the following section. Their analysis on multi-view classification utilizes datasets with complete data; here we seek to extend their work by answering a further question critical to real-world, real-time tasks: can models and learning paradigms generalize to cases of multiple data sources when significant data is missing? 

This problem of missing, corrupted, or asynchronized multi-modal data is found in many domains, ranging from biomedical imaging modalities like photoacoustic and computed tomography and optical microscopy \cite{wang2021advances} to autonomous systems dealing with temporally-calibrated LiDAR, vision, and radar \cite{yeong2021sensor} and identification of crop disease from satellite imagery with significantly different capture frequencies or resolutions \cite{ouhami2021computer}. 

In our analysis, we examine ``best-of-$N$" performance from collections of $N$ independent models, as well as schemes which negotiate between the logits of $N$ independent models, and a model which learns between hidden features derived from the $N$ data sources jointly, known as \textit{late fusion}. We adopt the term ``ensemble" to refer to the $N$ models respectively learned from the $N$ data sources, which may be combined to generate a prediction.  Critically, under our condition of irregular redundancy, the number of views available varies between instances, thus requiring the introduction of our method for multi-view ensemble learning with missing data. Further, because there are multiple simultaneous tasks involved in driver monitoring, we examine the \textit{task relevance} of each modality \cite{liang2022foundations} in our analysis. 

To summarize our contributions, we (1) perform comparative analysis between single-view, ensemble voting-based, and late-fusion learning on data from four real-world, continuous-estimation safety tasks, using sensors operating with irregular redundancy, (2) provide a generalized formulation of the real-world, real-time multi-modal problem such that our methods can be applied to similar tasks in both autonomous driving and other domains, and (3) evaluate the performance of these models with respect to human-centric safety systems by examining task performance on human drivers outside of the training datasets.

\begin{table}[]
    \caption{Multiview Fusion Methods. Homogeneity (from \cite{liang2022foundations}) refers to the extent that the abstract information presented in one view is equivalent to the information presented in another, toward the intended task(s). High homogeneity is highly redundant, while medium homogeneity refers to cases where some combinations of views may have the same information, but this information may be exluded from other views. Low homogeneity refers to situations where information between views is primarily supplemental. Our presented method is notable in having only medium homogeniety in support of its task, and frequent appearance of incomplete sets.}
    \label{tab:vismethods}
    \centering
    \small{
    \begin{tabular}{c|c|c|c|c}
       Method  &  Modalities & Tasks & Homogeneity & Incompleteness \\
       \hline 
       Various Fusions \cite{seeland2021multi} & 2-5 RGB & 1 & High & None \\ 
       Late Fusion \cite{silva2021multi} & 4 IR + 1 RGB & 1 & High & None \\
       Temporal Score Fusion \cite{negrete2023multi} & 3 RGB & 1 & High & None \\ 
       Late Fusion \cite{khajwal2023post} & 5 RGB & 1 & Medium & None \\
       Slice Fusion \cite{wu2022ensemble} & Depth Slices & 1 & Low & None \\
        \textit{\textbf{Ours}} & 4 IR & 4 & Medium & Frequent \\
    \end{tabular}
    }
\end{table}

\section{Related Research}

    \subsection{Sensor Fusion}
    
    Sensor fusion describes integration of data from multiple sensor sources, like LiDAR or cameras, towards a task. In the intelligent vehicles domain, research in methods of combination of output from multiple sensors to improve tasks in prediction and estimation is well-established. Chen et al. designed a Multi-Vew 3D Network (MV3D) that fuses LiDAR point cloud and RGB image data to perform 3D Object Detection in autonomous driving scnearios \cite{chen2017multi}. Their deep fusion of camera and LiDAR data uses FractalNet, a CNN architecture that is an alternative to other state-of-the-art CNNs like ResNet \cite{larsson2016fractalnet}. Similarly, Liang et al. fuse LiDAR and image feature maps into using a continuous convolution fusion layer \cite{liang2018deep}. This fusion process creates a birds-eye-view (BEV) feature map that is fed into a 3D Object Detection Model. Pointpainting is another prominent example of a fusion process of LiDAR and image data \cite{vora2020pointpainting}. Pointpainting takes image data and performs semantic segmentation to compactly summarize the features of the image. To fuse the LiDAR and image data, the LiDAR data is projected onto the semantic segmentation output.  In all these methods, the sensors are LiDARs and cameras. There is a different class of work which aims to do sensor fusion using the same modality or sensor type, but with data collected from different sensors or sensor views, like \cite{zhou2020end}, which combines LiDAR point clouds from the birds-eye-view and perspective view to learn fused features. In our work, we learn image features fused from different camera views. The features can be combined at different stages in the network giving rise to different ensembles, as explained in the next section. 

    \subsection{Ensemble Learning}
    
    In addition to fusion of output from $N$ sensors to reduce uncertainty of the observed information, we also explore ways that the models learned from the input of these $N$ sensors can share information during the learning process, such that the collective ensemble is optimized to the task. 
    
    We present here the Sagi and Rokach's survey definition of Ensemble Learning: 
    
    
    \begin{displayquote}
    ``Ensemble learning is an umbrella term for methods that combine multiple inducers to make a decision,...The main premise of ensemble learning is that by combining multiple models, the errors of a single inducer will likely be compensated by other inducers, and as a result, the overall prediction performance of the ensemble would be better than that of a single inducer." \cite{sagi2018ensemble}
    \end{displayquote}
    
    It is worth noting that there are a variety of methods for generating such ensembles \cite{sagi2018ensemble}. For example, the high-level learning system may:
    \begin{enumerate}
        \item Vary the training data provided to the inducers \cite{chan1995comparative} \cite{chawla2004learning} \cite{rokach2008genetic} \cite{ting2011feature},
        \item Vary the model architecture between inducers \cite{wen2017ensemble} \cite{ayad2010voting} \cite{deng2014ensemble},
        \item Vary the learning methodology \cite{brown2005managing} or hyper-parameters \cite{lin2012parameter} between inducers, or 
        \item Vary some combination of the above between inducers. 
    \end{enumerate}
    
    In this research, we freeze the model architecture, learning methodology, and hyper-parameters; we vary only the training data provided to the inducers. However, in this case, the training data is not sampled or refactored from some shared pool; instead, each ensemble member has access to its own set of training data. These training data are not independent, though; the training data is unified as a \textit{collection} per instance, where each member of the collection is a different representation of the same base observation (e.g. different cameras taking simultaneous photos of the same object). 
    
    From the ensemble of inducers, inductions can be combined and learned-from in a variety of ways:
    
        \subsubsection{Bayesian model averaging and combination}
        Bayesian Model Averaging (BMA) allows formation of predictions with many candidate models without losing information like an all-or-none technique. Using Bayesian model averaging, the probability of a prediction $y$ given training data $D$ can be defined as:
        \begin{equation}
        p(y | D) = \sum_{k=1}^K p(f_k | D)p(y | f_k, D)
        \end{equation}
        where $f_k$ is the prediction of the kth model. The posterior probabilities $p(f_k | D)$ can be treated as weights $w_k$ for each of the separate models since $\sum_{k=1}^Kp(f_k | D) = 1$. Previously researched applications using these methods include weather forecasting \cite{raftery2005using}, flood insurance rate maps \cite{huang2023uncertainty}, risk assessment in debris flow \cite{tian2023data}, and crop management \cite{fei2023bayesian}. 
        
        As mentioned by Monteith et al., Bayesian Model Averaging can be more thought of as a model selection algorithm, as ultimately the importance of each model is determined by the posterior probability weight \cite{monteith2011turning}. To develop an approach that is more inherent in ensemble learning, there are various strategies that can be used for model combination rather than model averaging. Bayesian model combination has found success in reinforcement learning by combining multiple expert models \cite{gimelfarb2018reinforcement}, speech recognition \cite{sankar2005bayesian}, and other tasks (notably functioning on non-probabilistic models and combinations of models observing different datasets \cite{kim2012bayesian}). 
    
    
    \subsubsection{Voting, Weighted Majority Algorithm}
    In ensemble learning, it is crucial to learn the value of each individual model by assigning different weights to these models in order to increase performance. In this algorithm, weighted votes are collected from each model of the ensemble. Then, a class prediction is made based on which prediction has the highest vote. All models which made incorrect predictions will be discounted by a factor $\beta$ where $ 0 < \beta < 1$ \cite{littlestone1994weighted}. Weighted majority algorithms have been used to combine model predictions to identify power quality disturbances for a hydrogen energy-based microgrid \cite{bayrak2022deep}, calendar scheduling \cite{blum1997empirical}, profitability and pricing \cite{braouezec2010committee}, and other applications. The weighted majority algorithm is used to combine the predictions from the expert models and see how effective each expert model is.

    In addition to single-view results, we compare performance in our hand classification task under naive voting, Bayesian model combination, and weighted majority voting. 
    

    \subsection{Machine Learning from Multiple Cameras}
    
\begin{figure}
        \centering
        \includegraphics[width=.75\textwidth]{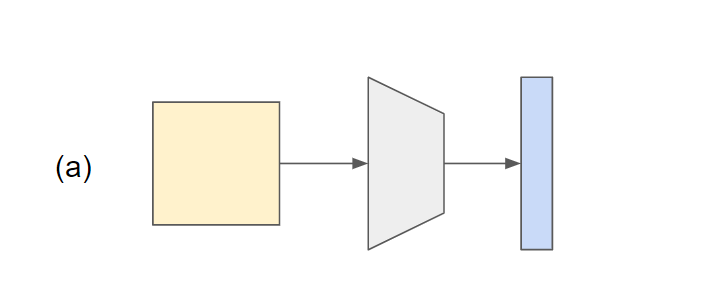}
        \includegraphics[width=.75\textwidth]{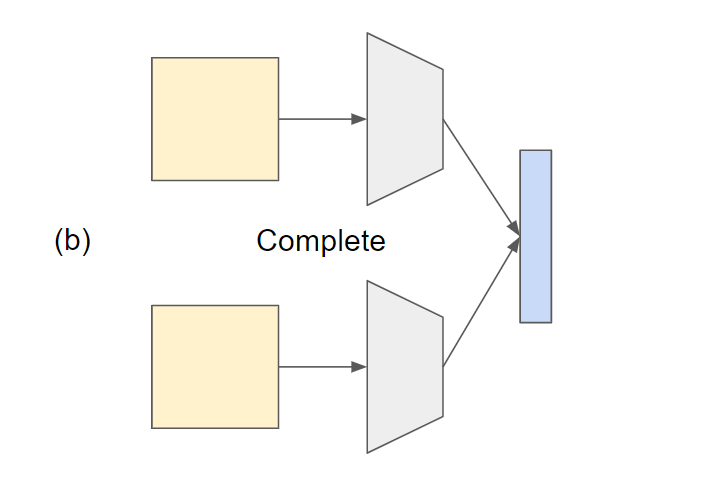}
        \includegraphics[width=.75\textwidth]{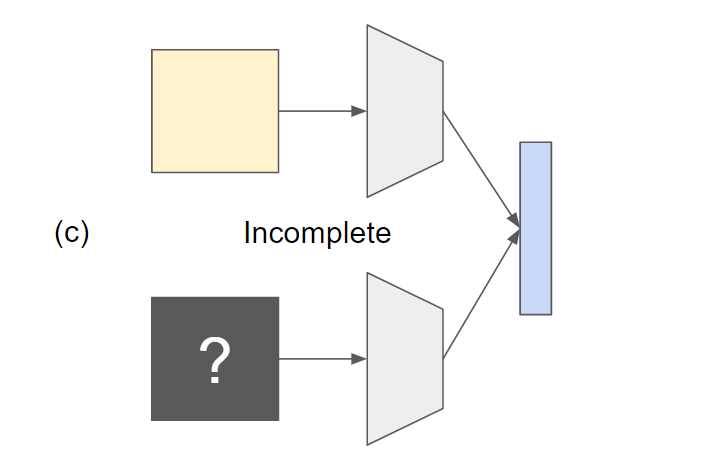}
        \caption{While traditional learning problems (a) may seek to learn a model (gray) which makes a prediction (blue) from input (yellow), in the multi-modal setting (b), we seek to learn a model which makes a prediction from multiple inputs. However, in the case where a sensor fails, becomes occluded, or operates at a different rate, the input set goes from \textit{complete} to \textit{incomplete}. In this research, we explore techniques for dealing with such incomplete sets (c), important for systems which are relied upon for always-online output prediction.}
        \label{fig:framing}
    \end{figure}

    Seeland and Mäder thoroughly investigate image classification performance gains afforded by network fusion at different process levels (early, late, and score-based) when using multiple views of an object \cite{seeland2021multi}. They apply their methodology to datasets comprising cars (shot from 5 views), plants (shot from 2 views), and ants (shot from 3 views). They find that late fusion provides the strongest performance gain for the car and plant datasets, and that an early fusion (slightly misnamed in this case, as it occurs at the final convolution) leads to a very marginal gain compared to late fusion on the ant dataset. In general, the authors results support late fusion as the dominant methodology, with early fusion often leading to worse performance compared to baseline. In their research, each data instance is referred to as a ``collection" (i.e. collection of $N$ images). Critically, each collection analyzed is \textit{complete}; that is, no view is missing from any given collection. This is where our problem framework and approach differs; as illustrated in Figure \ref{fig:framing}, we consider situations in which collections may be incomplete, and seek to learn correct labels despite missing data.

    The late fusion approach for visual patterns has found success in multiple application domains, as presented in Table \ref{tab:vismethods}, and even in other domains such as cross-modal information retrieval \cite{liang2022multiviz} \cite{beltran2021deep} \cite{wan2022editorial} \cite{xue2023dynamic}.

    \subsection{Driver Hand Activity Classification}
    
        \subsubsection{Shortcomings of Non-Camera Methods}
        At the time of writing, most commercial in-vehicle systems which monitor the driver’s hands use pressure and torque sensors embedded in the steering wheel to detect the presence of a driver's hands. However, this method of sensing leaves multiple safety vulnerabilities: 
        \begin{enumerate}
            \item Especially in the case of non-capacitative sensing, these sensors can be spoofed by placing weighted objects on the wheel, leading to recent fatal accidents.
            \item When effective for determining if the hands are on or off the wheel, these sensors still cannot distinguish between different hand activities taking place off of the wheel, and recognizing these activities is critical for estimating important metrics like driver readiness and take-over time. Hand locations and held objects imply hand activities, crucial to inferring a driver’s state, and this information is lost when reduced to hands-on-wheel and hands-off-wheel.
        \end{enumerate}
    
        \subsubsection{Camera Methods}
    
        Camera-based methods of driver hand analysis allow for observation of the hands without steering wheel engagement. Past systems for classification of driver activity and identification of driver distraction use traditional machine learning approaches; for instance, Ohn-Bar et al. demonstrated systems that utilize both static and dynamic hand activity cues in order to classify activity in one of three regions \cite{ohn2013driver} and extracts various hand cues in ROI and fuses them using an SVM classifier \cite{ohn2014head}. Borgi et al. use infrared steering wheel images to detect hands using a histogram-based algorithm \cite{8373883}. More recent works expand on the aforementioned classifiers and utilize deep learning in order to identify and classify driver distraction in a more robust manner. Eraqi et al., among others, have developed systems that operate in real time to identify driver distraction in a CNN-based localization method \cite{eraqi2019driver}. Shahverdy et al. also use a CNN-based system in order to differentiate between driving styles (normal, aggressive, etc.) in order to alert the driver accordingly \cite{shahverdy2020driver}. Building on this, Weyers et al. demonstrate a system for driver activity recognition based on analysis of key body points of the driver and a recurrent neural network \cite{weyers2019action}, and Yang et al. further demonstrate a spatial and temporal-stream based CNN to classify a driver’s activity and the object/device causing driver distraction \cite{yang2020refined}. A comprehensive survey outlining the current driver behavior analysis using in-vehicle cameras was done by Wang et al \cite{wang2021survey}.
        
        Recent pose detection models provide another helpful tool in understanding the hands of the driver. As defined by Dang et al., 2D pose detection involves detecting important human body parts from images or videos \cite{dang2019deep}. Chen et al. describes that there are three ways to define human poses: skeleton-based models, contour-based models, and 3D-based volume models \cite{chen20222d}. Our research uses a skeleton based-model, which describes the human body by identifying locations of joints of the body through 2D-coordinates. A deep learning approach to pose detection through a skeleton based-model is to first detect the human location through object detection models like Faster-RCNN and then perform pose estimation on a cropped version of the human. Some successful approaches to pose detection include HRNet, which is successful for pose detection problems since it maintains high-resolution representations of the input image throughout a deep convolutional neural network \cite{sun2019deep}. Toshev and Szegedy perform this pose estimation by implementing the model DeepPose, which refines initial joint predictions via a Deep Neural Network regressor using higher resolution sub images \cite{toshev2014deeppose}. Yang et al. design a Pyramid Residual Model for pose estimation which learns convolutional filters on various scales from input features \cite{yang2017learning}. 
        
        Though consumer and commercial vehicles have begun integrating inside-facing cameras for a variety of tasks, such as attention monitoring and distraction alerts, these methods are not without their own challenges. A single camera may be well-suited to a particular task, but different situations may call for different camera placements. While one view may be ideal for a particular task within design constraints, this view may sacrifice a complete view of a different driver aspect and may not offer redundancies if a camera is obstructed or blocked. For example, an ideal hand view (taken from above the driver) would not be suitable for assessing a driver’s eyegaze, but a camera that can see the driver’s eyes may also have at least a partial view of the driver’s hands.

    \subsubsection{Safety and Advanced Driver Assistance Systems}
\begin{figure}
        \centering
        \includegraphics[width=\textwidth]{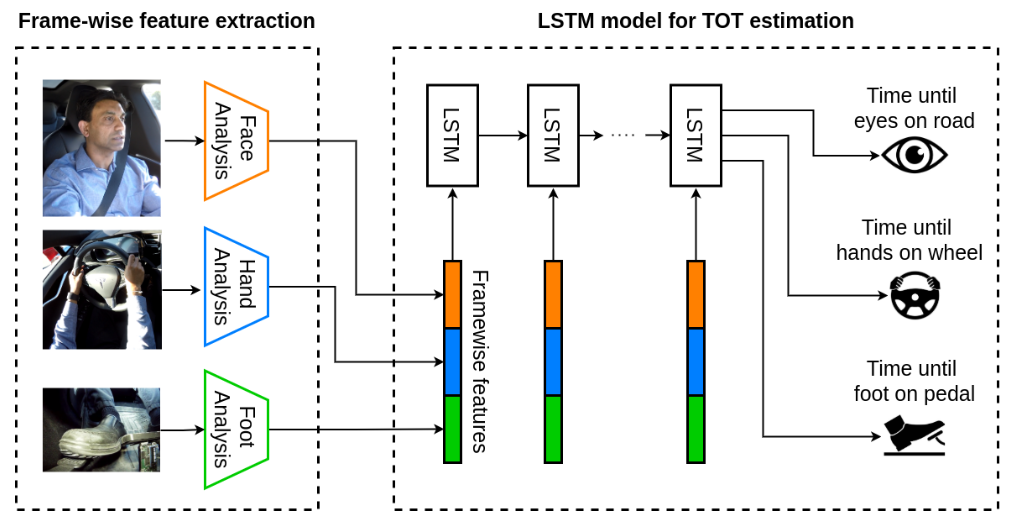}
        \caption{As illustrated in \cite{rangesh2021predicting}, analyzing logits of hand activity and location classes play a useful role in predicting a driver's readiness to take control of a vehicle.}
        \label{fig:hand_help}
    \end{figure}

\begin{figure}
        \centering
        \includegraphics[width=\textwidth]{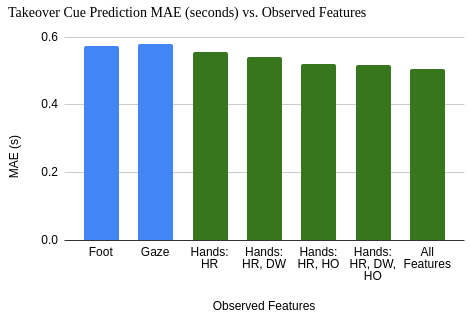}
        \caption{In an ablative study, Rangesh et al. \cite{rangesh2021autonomous} show that various individual features and combinations of features associated with the hands, including hand region (HR), distance to wheel (DW), and held object (HO) are most informative to models for predicting cues associated with vehicle takeovers from automated to manual control. In the case of control transitions, these fractional-second gains are critical for a driver's reflexes to safety alerts.}
        \label{fig:handimportance}
    \end{figure}

    Recent works in safety and advanced driver assistance systems utilize deep learning techniques in order to perform driver analysis. In particular, deep learning allows researchers to extract driver state information and determine if they are distracted through analyzing driver characteristics such as eye-gaze, hand activity, or posture \cite{wang2021survey}. Estimating driver readiness is another vital aspect to safe partial autonomy, and a key component to understanding driver readiness is hand activity, as a distracted driver often has their hands off the wheel or on other devices like a phone. Illustrated in Figure \ref{fig:handimportance}, Rangesh et al. \cite{rangesh2021autonomous} and Deo \& Trivedi \cite{deo2019looking} show that driver hand activity is the most important component of models for prediction of driver readiness and takeover time, two metrics critical to safe control transitions in autonomous vehicles \cite{greer2023safe} \cite{cummings2021safety}. Such driver-monitoring models take hand activity classes and held-object classes as input, among other components, as illustrated in Figure \ref{fig:hand_help}. These classes can be inferred from models such as HandyNet \cite{rangesh2018handynet} and Part Affinity Fields \cite{yuen2019looking}, using individual frames of a single camera view as input. Critically, this view is taken to be above the driver, centered in the cabin and directed towards the lap– a typically unobstructed view of the hands. 

    Application of multi-view and multi-modal learning to safe, intelligent vehicles (\cite{roitberg2022comparative}, \cite{tawari2014robust}) brings two benefits: increased flexibility in field-of-view for individual component cameras, and increased accuracy in classification for observable activity. Both benefits arise from the ability of the system to reason between views, allowing occluded or otherwise compromised images from one view to be substantiated by images from additional views in cooperation.

\section{Methods}

    \begin{figure}
        \centering
        \includegraphics[width=\textwidth]{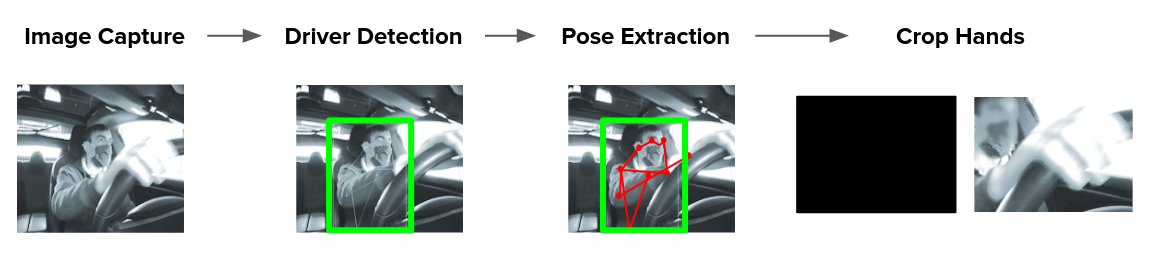}
        \caption{The image preprocessing pipeline prior to learning involves four steps, carried out individually from each camera stream. First, the image is captured, then, the driver is detected and their pose extracted, allowing for crops around the hands to be generated. In this example, because the left hand is not visible to the particular camera, the method of single imputation is used to replace the frame with a frame of zeros. We note that because the method uses only the image of the hands towards its learning, it is possible to anonymize the driver by blurring the face, as we have done in the above example, for the cropped frames that serve as model input.}
        \label{fig:preprocess}
    \end{figure}

    \begin{figure*}[h]
        \centering
        \includegraphics[width=.99\textwidth]{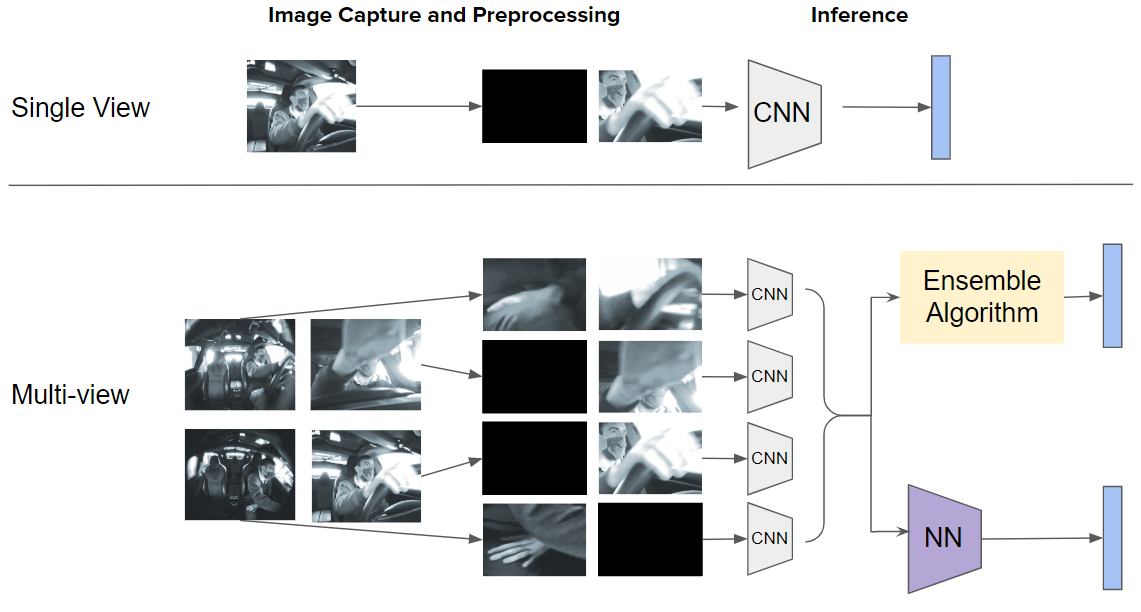}
        \caption{Classification pipeline. Following image capture, we perform image processing to detect the driver using Faster-RCNN with Feature Pyramid Networks (FPN) with a ResNet-50 backbone, extract the driver pose using HR-Net, and crop the hands 100px from center of wrist joints. In the Inference stage, we utilize CNNs for classification, beginning from a pre-trained ResNet fine-tuned on our dataset. For the single view model, we make direct inference, and for the multi-view models, we pass the logits to ensemble algorithms, or pass the CNN-output feature maps to a neural network for late fusion. In our experiments, we use Bayesian Combination and Weighted Majority Averaging as the Ensemble Learning algorithms, and Late Fusion via fully-connected neural network laters.}
        \label{fig:arch}
    \end{figure*}

    The general hand activity inference stage is organized in four steps: multi-view capture, pose extraction, hand cropping, and CNN-based classification. 
    

    \subsection{Pre-processing Steps}
        \subsubsection{Feature Extraction: Pose and Hands}
\begin{figure}
            \centering
            \includegraphics[width=\textwidth]{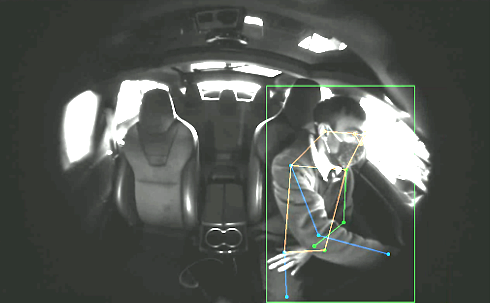}
            \caption{Prior to classifying driver hand activity, the system must detect the driver. We use Faster-RCNN to generate the bounding box shown in green. Following driver detection, we apply HRNet to identify the 2D pose skeleton, shown as keypoints and connecting lines on the driver's body.}
            \label{skeleton}
        \end{figure}

            The inference stage is illustrated in Figure \ref{fig:preprocess}. Following data capture, we extract the pose of the driver in each frame, where “pose” is a collection of 2D keypoint coordinates associated with the driver’s body, such as the wrists, elbows, shoulders, eyes, etc.  This problem is broken into two steps: first, we must detect the driver in the frame, then detect the driver’s pose. Each step requires its own neural network; for driver detection, we first use the Faster-RCNN \cite{girshick2015fast} model with Feature Pyramid Networks \cite{lin2017feature}, using a ResNet-50 backbone \cite{He2015DeepRL} to detect the driver. We note that this network will output any humans detected in the frame, so we apply a post processing step (based on the camera view) to only include detections corresponding to the driver’s seat. For joint detection, we employ the HRNet \cite{wang2019deep}, a robust top-down pose detection model,which predicts 2D coordinates of various points of the body such as the wrists, elbows, shoulders, eyes, etc. The results of driver and keypoint detection are illustrated in Figure \ref{skeleton}. 
            
        \subsubsection{Hyperparameter Selection: Hand Crop Dimensions}
            We crop images around each of the hands, centered at the wrist and extending 100 pixels in each direction. The width of the crop is a hyperparameter which can be changed to add or reduce spatial context. Only these hand crops are fed into the activity classification pipelines.

    \subsection{Single-View Models}
    The cropped images from a particular camera are classified by two convolutional neural networks trained on images of that view. One network outputs probabilities that the hands are holding one of three objects: Phone, Beverage, Tablet; or holding nothing. The second network (identical in architecture to the first, except for number of classes) predicts the probability that the hand is in one of five hand location classes: Steering Wheel, Lap, Air, Radio, or Cupholder. The classes Radio and Cupholder are reserved for the right hand only. For single-view model evaluation, the network infers the hands to be classified according to the class of maximal probability. 
    
    In cases where there is no image available, the model is provided an image of proper dimension containing only the value 0. This is a variation of the method referred to as single-imputation \cite{wang2021advances}, in which a single value is used to replace any instances of missing data. The intention behind this decision is that the network will learn a prior over the training data in situations when the view is occluded; that is, each time a blank image is presented, it infers that the sample should be classified in one of the typically occluded positions, with probability representative of the distribution of the training data. 
    
    \subsection{Naive Voting}
    In the naive voting scheme, all four single-view models make a prediction using their respective image from a given collection, noting that up to $N-1$ images may be blank. The prediction made by the network is taken to be
        \begin{equation}
            y =  \arg \max_{i}(\sum_{j=1}^{N}\frac{1}{N} p_{ij}),
        \end{equation}
    where $M$ is the number of classes, $N$ the number of models, and $p_{ij}$ the probability of the $i$th class from the $j$th model. This method gives each model equal vote. 
    
    \subsection{Weighted Majority Voting}
    Using Weighted Majority Voting, we seek to combine the decisions of the 4 models weighted by a discount factor $d_i$. This discount factor is based on the number of mistakes $m_i$ made by the model during validation:
    \begin{equation}
    d_i = 1 - \frac{m_i}{\sum\limits_{i=1}^{N}m_i}
    \end{equation}
    
    Then, each collection prediction is made using
    \begin{equation}
        y = \arg \max_{i}(\sum_{i=1}^{N}d_i p_i). 
    \end{equation}

    \subsection{Bayesian Model Combination}
    Using Bayesian Model Combination, we combine the decisions of the 4 models weighted by a factor representing the likelihood of the particular model given the observed data, $P_i \sim p(f_i|D)$. In cases where the hands are not detected in a certain view $i$, then we consider model $f_i$ to have low likelihood; therefore, we set $P_i$ to zero in such situations. If $n$ models have $P_i$ as zero, then the $P_i$ of remaining models is distributed uniformly as
    \begin{equation}
        P_i = \frac{1}{N-n}
    \end{equation}
    where $N$ is total number of views.
    \begin{equation}
        y = \arg \max_{i}(\sum_{i=1}^{N}P_i p_i)
    \end{equation}
    
     \subsection{Multi-view Late Fusion}
    
     For the late fusion scheme, we use a neural network architecture composed of four parallel sets of convolutional layers (ResNet-50 backbones), which act on each of the four image views. Following the convolutional layers, each parallel track is fed to its own fully-connected layer of 512 nodes (followed by a ReLU activation). These layers are joined together by a fully-connected layer with 2048 nodes (followed by a softmax activation); this is the point of fusion, where the features extracted from the four views are combined and the relationships between the multiple views are learned. 
     
     We call this late fusion as it is done at the penultimate layer, late in the pipeline. This was done to make sure the fusion could leverage high level features present deeper in the pipeline. We use two fused models as before; one which outputs probabilities that the hand is holding one of the 3 objects and another which outputs the location of the hand. The maximal probability class is chosen as the classification output.

\section{Experimental Evaluation}

\subsection{Comparison of Single-View and Ensemble Techniques}
Using four cameras, we collect a dataset of 19 subjects engaged in various hand placements and object-related activities. 

\begin{figure}[h]
\centering
     \includegraphics[width=0.49\textwidth]{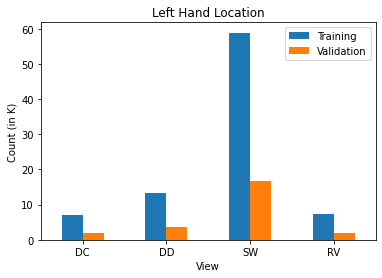}
    \includegraphics[width=0.49\textwidth]{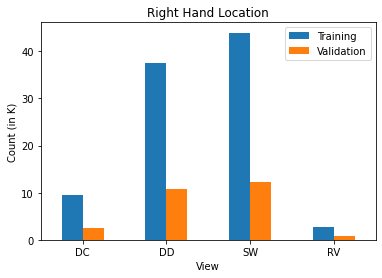}
       \includegraphics[width=0.49\textwidth]{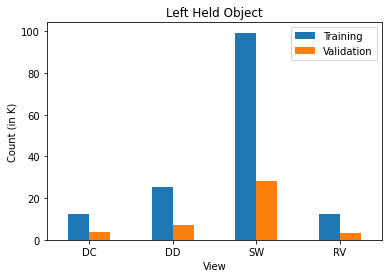}
        \includegraphics[width=0.49\textwidth]{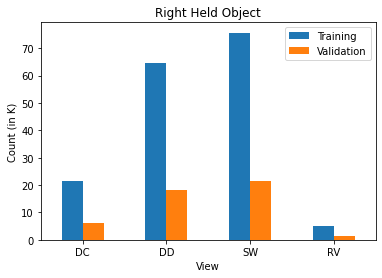}
    \caption{Distribution of collected samples for locations (Left and Right hand) and held objects (Left and Right hand)
    from 4 views Dashboard Driver (DD), Dashboard Center (DC), Steering Wheel (SW) and Rear View (RV)}
    \label{fig:dist}
\end{figure}

Altogether, we collect approximately 81,000 frames corresponding to hand zone activity, and 128,000 frames corresponding to held object activity. We divide these into training, validation, and test sets using approximately 80\%, 10\%, and 10\% of the data respectively (with marginal differences to account for dropped frames). The distribution of the data between views is shown in Figure \ref{fig:dist}. We note that the challenges of selecting camera views for this task are readily apparent in the proportions of the data; one camera view (rearview) has significantly less frames where the pose is reliably estimated, while the steering wheel view has many. However, the availability of frames does not necessarily correspond to the ability of that view to be informative to the task at hand nor generalizability to other tasks in the autonomous driving domain. 

Using this data, we trained the above-described neural networks for hand location and held object classification into the defined zones (3 location zones for the left hand, 5 location zones for the right hand, and 4 held objects [including null] for each hand).

\subsubsection{Single-View Models}

We evaluate the four single view models on images from the test set, including blank images when no image is available. From this, we compute an average model accuracy by taking the average of each per-class accuracy for each of the four tasks (left hand location, right hand location, left hand held object, right hand held object). For each task, we report the performance of the best-performing model, the worst-performing model, and the average across the four models. This highlights foremost the importance of camera view selection for this particular task, but also provides a point of comparison to see how the ensemble learning and fusion methods may enhance the overall performance of the models to their task. Results are provided in Table \ref{acc-table}.

\begin{table*}
\caption{Classification accuracies (averaged across all classes) of baseline single-camera views. The rows represent, for each task, the performance of the best-performing and worst-performing of the $N$ camera view models, as well as the average performance across views.}
\label{acc-table}
\vskip 0.15in
\begin{center}
\begin{tabular}{||c c c c c ||} 
 \hline
 Method &  LH Location & RH Location & LH Held Object & RH Held Object \\ 
 \hline\hline
 Worst-of-N & 0.442 & 0.212 & 0.322 & 0.289 \\ 
 \hline
 Average-of-N & 0.593 & 0.458 & 0.513 & 0.581 \\
 \hline
 Best-of-N & 0.952 & 0.785 & 0.952 & 0.836 \\
 \hline
\end{tabular}
\end{center}
\end{table*}

\subsubsection{Ensemble Methods: Naive Voting, Weighted Majority Voting, Bayesian Model Combination, and Multi-View Late Fusion}

We evaluate the four methods described in the Methods section, as well as an additional method which employs both Weighted Majority Voting and Bayesian Model Combination simultaneously. We evaluate the performance of these models on two different sets: first, only on collections which have all $N$ images available, and second, on collections with any number of images (1 to $N$) available. Results are provided in Tables \ref{acc-table2} and \ref{acc-table3}.

Importantly, only a very small fraction (less than 3\%) of each of our task test set collections are complete, as shown in Table \ref{testsize}. In fact, some task classes are never simultaneously observed from all views, so results in Table \ref{acc-table2} indicate performance on a limited number of classes from the actual task at hand, and at that, only for complete collections! While the models may be great at making inference when they have a clear view of the object of interest, this suggests a significant performance gap for a safety system expected to make continuous inference across all classes, not just inference when data is complete. By contrast, Table \ref{acc-table3} represents performance across \textit{every} sample of the test set. We include both tables to illustrate the point that while the voting-based methods begin to fail, the late fusion method performs just as well even when data is missing from a collection.

\begin{table}
\caption{Test set size for different tasks, and percentage of test set collections which are complete.}
\label{testsize}
\vskip 0.15in
\begin{center}
\begin{tabular}{||c c c||} 
 \hline
 Task &  Test Set Size & \% Complete \\ 
 \hline\hline
 LH Location & 9,193 & 2.43\% \\
 \hline
 RH Location & 9,205 & 1.67\% \\
 \hline
 LH Held Object & 15,486  & 2.59\% \\ 
 \hline
 RH Held Object & 14,624 & 2.22\%  \\
 \hline
\end{tabular}
\end{center}
\end{table}

\begin{table*}
\caption{Classification accuracies (averaged across all classes) of different ensemble methods on four hand classification tasks, evaluated only when all $N$ views are available. In this ``complete-view-only" test set, 2 classes from left hand location, 4 from right hand location, and 1 each from left and right hand held object are completely unrepresented. Performance on the held object tasks may be poor due to the uncertainty in less-informative views bringing down the overall confidence of the system towards the correct class (or artificially raising confidence in the incorrect class). Naive voting may outperform weighted majority voting when challenging examples found in the validation set may be unrepresented in this test set, thereby discounting models which would otherwise be ``correct". This table also serves to illustrate how often frames are missing in these tasks, demonstrating the importance of a method which is robust to missing data.}
\label{acc-table2}
\vskip 0.15in
\begin{center}
\begin{tabular}{||c c c c c ||} 
 \hline
 Method &  LH Location & RH Location & LH Object & RH Object \\
 \hline\hline
 Naive Voting & 1.000 & 0.981 & 0.509 & 0.426 \\
 \hline
 \small{Weighted Majority Voting} & 0.991 & 0.987 & 0.403 & 0.410 \\
 \hline
 Late Fusion & 1.000  & 1.000 & 0.978 & 0.995 \\ 
 \hline
\end{tabular}
\end{center}
\end{table*}

Our original question was: can these methods overcome situations where data is missing from a collection? Table \ref{acc-table3} provides our answer. When data is missing, the voting-based methods struggle significantly due to the falsely-placed confidence given to the model output. The largest challenge with these approaches is recognizing which view is dominantly correct in a particular situation and leveraging that view appropriately; otherwise, too much weight may be given to a model which has false confidence, and a model's vote may be a reflection of the intrinsic difficulty of that particular view. Able to better leverage information between views, the best performance comes from the multi-view late fusion approach. The late fusion model both (1) maintains near-perfect performance on the four tasks, even when 1 to $N-1$ frames are missing from the collection, and (2) exceeds performance of all single-view cameras for each task. These two results suggest that a late-fusion model is successfully learning complementary information that is unavailable in a single-view; that is, the model is effective in combining different sources of information to make a better-informed prediction on the task. Additionally, it is able to do so despite missing data, suggesting that the model has learned to leverage remaining sources of information when frames are dropped. 

\begin{table*}
\caption{Classification accuracies (averaged across all classes) of different ensemble methods on four hand classification tasks, with 1 to $N$ views available in each collection.}
\label{acc-table3}
\vskip 0.15in
\begin{center}
\begin{tabular}{||c c c c c ||} 
 \hline
 Method &  LH Location & RH Location & LH Object & RH Object \\
 \hline\hline
 Naive Voting & 0.470 & 0.205 & 0.277 & 0.334 \\
 \hline
 \small{Weighted Majority Voting} & 0.443 & 0.201 & 0.269 & 0.316 \\
 \hline
 \small{Bayesian Model Combination} & 0.397 & 0.338 & 0.366 & 0.360 \\
 \hline
 WMV+BMC & 0.398 & 0.338 & 0.358 & 0.363 \\
 \hline
 Late Fusion & \textbf{0.991} & \textbf{0.988} & \textbf{0.978} & \textbf{0.986} \\ 
 \hline
\end{tabular}
\end{center}
\end{table*}

In prior work, Greer et al. \cite{greer2023multiview} show that multi-view late fusion models give superior results over single-view models because the network can learn from more perspectives. Late fusion is particularly effective as all the camera views have high-level richer features deeper in the pipeline. The multi-view late fusion model was successful in classifying zones and objects when the training and test subjects were same. But in real-world scenarios, models need to generalize to unseen subjects. We elaborate on our approach for evaluating performance on cross-subject  classifications in the next section, and provide recommendations for such systems in the following discussion.

\subsection{Multiple Subject Validation: Generalizing to Unseen Drivers}

In the first set of experiments, we show that multi-view late fusion models give superior results over single-view models because the network can learn from more perspectives. Late fusion is particularly effective as all the camera views have high-level richer features deeper in the pipeline. The multi-view late fusion model was successful in classifying zones and objects when the training and test subjects were same. But in real-world scenarios, models need to generalize to unseen subjects. Here, we evaluate performance on cross-subject  classifications, and provide recommendations for such systems in the following discussion.

Greer et al. \cite{greer2023multiview} evaluate the late-fusion model performance on a substantial set of test data derived from the same capture system and subjects as the training and validation data, but in intelligent vehicle applications, it may be impractical to collect training data on each individual driver. An ideal model would generalize to all drivers that may use the vehicle. 

A typical risk in end-to-end learning on overparameterized systems involving human subjects is that such a deep neural network is not typically ``explainable" \cite{sachdeva2023rank2tell} \cite{liang2022multiviz}. When the model learns from humans, it can overfit to particular features associated with an individual subject, rather than learning actual patterns of interest (e.g. the model becomes really good at learning how to recognize Subject A's hand holding Subject A's cell phone, rather than a more general prototype of any hand holding any cell phone).

Machine learning models are commonly evaluated using k-fold cross validation, but this evaluation has shortcomings when data from the same subjects are contained in both train and validation sets, since (as described above) the model can overfit to the subject's unique signature instead of the latent activity. Accordingly, techniques of subject cross validation are preferred \cite{dehghani2019subject}. In typical k-fold cross validation, data is divided into k sets, and each of these k sets have a turn being left out of the training process (used only for evaluation). The summary statistics to describe the goodness of the model is then the average model performance on the k validation sets. 
    
In our case, we utilize a dataset of 19 subjects. Here, we discuss evaluation choices made on splitting the data. We first constrain evaluation such that any subject being used in validation is unseen during training. We note that early stopping is controlled by a subset of the training data to prevent significant overfitting; while it may be beneficial to let yet another unseen subject (or subjects) determine the training stop-point, this introduces the bias of model performance to that particular driver (or drivers), which will not necessarily translate to performance on the unseen evaluation driver. 
    
We use varying values of $k$ on each task to handle computational constraints, rotating a left-out subject from each of the $k$ model trainings for each task. For each model, we take the average accuracy among all of the classification categories (the so-called \textit{macro-averaged precision}), and then average this value among the $k$ models. We evaluate using $k=8$ for the left and right hand location tasks, $k=17$ for the left hand held object task, and $k=13$ for the right hand held object task.

We report this averaged performance for each of the four single camera views as well as the late-fusion multiview model, in Table \ref{acc-table}. 

We first note that, with the exception of the rearview-mounted camera on two left hand tasks, single-view models do not seem to generalize well across subjects; classification on the unseen subject tends to collapse to a few classes, likely due to overfit to a nearest-neighbor image in the training data. Practically speaking, this would suggest that models for driver state estimation which rely on a single camera would indeed benefit from fine-tuning on data from the driver of interest; we know that the model can train to near-perfect accuracy on data it has seen, it's the generalizability that causes the issue. 

Now, to our primary question: can late-fusion multiview models overcome the generalizability challenge? Our results suggest that the late-fusion multiview model does outperform the best of the single-view models for unseen drivers on right-hand related tasks, though the rearview-mirror placed camera is excellent at left-hand related tasks. The late-fusion multiview model exceeds the average between the four single cameras on every task.  
    \begin{itemize}
        \item For the left hand location task, the late-fusion multiview model is 9.6\% less accurate than the best-performing rearview model, but 33\% more accurate than the average across camera views.  
        \item For the right hand location task, the late-fusion multiview model is 10.8\% more accurate than the best-performing rearview model, and 45\% more accurate than the average across camera views. 
        \item For the left hand held object task, the late-fusion multiview model is 6\% less accurate than the best-performing rearview model, but 30\% more accurate than the average across camera views. 
        \item For the right hand held object task, the late-fusion multiview model is 4.3\% more accurate than the best-performing dashboard-center-view model, and 15\% more accurate than the average across camera views. 
    \end{itemize}

\begin{table*}
\caption{Classification accuracies (averaged across all classes) of single-camera and late-fusion multiview models on four hand activity tasks. Evaluations are averaged over 19 models in which the evaluated subject is unseen during training, with mean and variance provided. View 1 is taken from the dashboard center, view 2 from the dashboard facing the driver, view 3 from the steering wheel, and view 4 from the rearview mirror. LFM refers to Late Fusion Multiview.}
\label{acc-table}
\vskip 0.15in
\begin{center}
\begin{tabular}{||c c c c c | c ||} 
 \hline
 View &  LH Location & RH Location & LH Object & RH Object & Average \\ 
 \hline\hline
 1 & 0.333, 0.001 & 0.229, 0.007 & 0.320, 0.004 & 0.442, 0.006 & 0.331 \\ 
 \hline
 2 & 0.217, 0.023 & 0.090, 0.008 & 0.276, 0.001 & 0.292, 0.004 & 0.219\\
 \hline
 3 & 0.317, 0.017 & 0.296, 0.027 & 0.259, 0.006 & 0.253, 0.08 & 0.281 \\
 \hline
4 & \textbf{0.861}, 0.024 & 0.664, 0.059 & \textbf{0.729}, 0.019 & 0.350, 0.014 & 0.651 \\
 \hline
 LFM & 0.765, 0.031 & \textbf{0.772}, 0.015 & 0.699, 0.023 & \textbf{0.485}, 0.049 & \textbf{0.680}\\
 \hline
\end{tabular}
\end{center}
\end{table*}

\section{Discussion and Concluding Remarks}
System designers often have interest in selecting optimal number (and placement) of sensors for a given task, and in this research, we explored methods of leveraging the irregular redundancy of multiple sensors observing the same scene. While one camera placed expertly may be sufficient at a single task (say, observing the hands), there are many other tasks relevant to safe driving, such as estimating eyegaze, passenger seating occupation and positioning, and distraction identification. 

What this framework contributes is a method of making stronger inference when information is missing from one source, and the system can recognize and leverage the fact the information is missing to then make better use of information in other sources. In fact, missing information often informs the other models; if a hand is not visible to one camera, then it is more likely to be within the view of another. Further considerations for enhanced accuracy include exploring weighting schemes for weighted majority voting, hyperparameter sweeps for crop sizes and model architecture, and model likelihood estimation for Bayesian model averaging. 

Late-fusion approaches which use our method of replacement of missing data with a zero-placeholder may effectively learn a prior distribution given a missed reading from a sensor or camera. This is particularly relevant in cases where multiple perspectives are necessary for complete observation, or when multimodal systems are used which sample at different rates. We see plenty of examples of this in existing technology; many phones and laptop computers use both RGB and IR cameras for securely identifying the user, and thermal cameras are often used as an additional modality for medical applications, but cameras operating on different spectra (or media) typically operate at different rates.

No camera perfectly captures an event, but by using ensemble learning and fusion, safety systems (where every inch of accuracy counts) may exploit the benefits in redundancy and completeness of multi-view or multi-modal observations. 
    \subsection{Efficient systems of multiple models}

        In this research, we evaluate performance of four models related to the driver's hands: two for held object (one for each hand), and two for hand location (one for each hand). While each model may function independently, cascading the models gives a better idea of a driver's current activity. For example, an image of the hand does not necessarily need to pass through both a held-object and hand-location model. In applications, the models can be cascaded such that first a held object can be determined, and if it is the case that no object is held, the image is passed forward to the location classification module. In fact, some applications may successfully ``short-circuit" for efficiency depending on their use; a left-hand holding a cell phone may be sufficient to send an advisory without necessary inference on its location nor the right hand's activity. 

        Of further note, the system bottleneck most strongly occurs at the level of 2D pose estimation. To review, the model first detects the driver, then estimates the driver's pose, and from this pose classifies smaller regions pertaining to the hands (or, for other applications, eyes and other keypoints of interest). Fortunately, a system will only need to pass through this bottleneck once per inference time, since the remaining downstream models all utilize the same predicted pose information (and are much less computationally expensive). Because this system is modular, continued research from the computer vision community on efficient 2D pose estimation will translate directly and smoothly to performance gains in such human driver analysis systems. 

        Further research may benefit from an analysis of the Vision Transformer architecture for this problem, since the Transformer is particularly adept at selecting which features should be attended to. However, the Transformer is notably computationally expensive, so any performance gains must be balanced with increased inference delays to meet application requirements. The application of attention maps for (near) human-explainable reasoning from multimodal streams is considered an open challenge within multimodal learning \cite{liang2022foundations}, and these sample tasks and irregularly redundant data sources may be a strong candidate for future experiments. 

    \subsection{Onboard vs. Cloud Processing}

    There is strong interest in moving compute from onboard processing toward cloud-based computing of driver monitoring data\footnote{\url{https://2023.ieee-iv.org/automotive-game-day}}, but of key concern for consumer support and adoption of such data schemes is the preservation of driver privacy. To this end, we highlight that our presented framework allows for extraction of particular features (rather than complete images), which then allows for the anonymization via blurring or pixel value adjustment of the driver's face or similar privacy-sensitive content before sharing toward network computers, since these components are unused in model training and inference.

    \subsection{System design recommendations from experimental results}

        Our results lead us to the following system design recommendations for applications involving camera-based driver state estimation:
        \begin{itemize}
            \item When possible, collect data and finetune models using the driver of interest. Generalizability is a difficult task since the real-world may violate the i.i.d. assumptions that allow for excellent performance from neural networks. Unseen data may not come from the same distribution as prior training; the simpler case is to fit the model to data that most closely matches the expect distribution (i.e. images of the intended driver). 
            \item If design constraints allow, opt for multiple cameras observing the driver to leverage complementary information between views, alternative views of occluded zones, and redundant information to provide improved accuracy and generalizability. 
            \item If restricted to a single camera view, an overhead view from a camera placed near the rearview mirror may be optimal. If unavailable, a view facing the driver from behind the steering wheel may provide the best performance on estimating whether the driver's hands are on the wheel or elsewhere. However, this view is less well-suited to infer what the driver is doing with their hands if off-wheel; for this, a camera view facing the cabin from the rearview area or dashboard is better suited. The selected view should be informed by the intended application use case. 
            \item While outside the scope of this research, we encourage applications implementing this framework to explore different hyperparameter values for crop size around the hands (or other features of interest). Differences in camera distance affect how much of the hand (and surrounding context) is visible within a particular crop size, and it may be worthwhile to vary these sizes for specific use cases depending on objects and locations of interest.
        \end{itemize}
    
    \subsection{Post-processing considerations for downstream applications}

        Systems that seek to reliably estimate the state of the driver's hands (or similar driver attributes) will have to apply robust thresholds and denoising techniques to distinguish between genuine distractions and momentary lapses in attention. 
        \subsubsection{Filtering} 
        We suggest low-pass filtering to reduce the effects of noisy patterns from inference (that is, small “blips” between classes for fractions of a second). This allows for a more steady prediction result by averaging over moving windows of time, where the window size is a hyperparameter that can be tuned based on observation of the duration of a typical inference mistake made by the network. 

        \subsubsection{Thresholding}
        We also suggest a thresholding step to distinguish between monetary lapses of attention (such as a driver quickly reaching for an object), versus an elongated period of distraction, which warrants an alert. The permissible interval of sustained distraction is another hyperparameter that should be tuned according to the goals of the automaker or driver policy. 

        \subsubsection{Alerting}
         If it is decided that the driver may be distracted, the system can then issue a standard request for driver attentiveness, or employ other downstream safety mechanisms. It is recommended that the alert system employs a method of alerting aligned to the standards of human-machine interface research; these techniques are outside the scope of this research, but we emphasize that this is a modular endpiece, and the presented framework can be applied for any downstream alerting mechanism. 

        \subsection{Additional applications}
        The hand activity framework requires multi-view capture, driver detection and pose estimation in its upstream steps. These tasks can be used towards several additional safety critical scenarios. As the driver detection step detects all individuals in the car, it can be used to estimate seat occupancy or passenger positioning. The cameras also capture driver gaze which can provide another signal towards driver attentiveness. The pose estimation module can provide data for studying safe airbag deployment in crashes. We believe demonstrating the effectiveness of multi-camera hand distraction can lead to further research in these applications to create holistic, robust, end-to-end systems for driver safety. 
\\

        There are many further layers of analysis to problems of irregular redundancy; in this research, we move beyond complete sets to emphasize approaches which are applicable toward incomplete sets. Future work should incorporate temporal dynamics into this analysis, towards making systems which show even stronger generality to new subjects. However, this temporal dependency differs from that described in \cite{liang2022foundations}, where the goal is ``to accumulate multimodal information across time so that long-range cross-modal interactions can be captured through storage and retrieval from memory" -- rather, we seek to retain short-range information from the collective representation, such that iterative predictions are consistent with prior predicted states. There is further promise in the ability of ensemble techniques to generalize to an entirely different (and relevant) class of what is ``unseen": in addition to generalizing to new subjects, domain-adaptive ensemble methods have also been shown to be effective learners to entirely new views \cite{zhou2021domain}, making them highly appropriate towards driver monitoring domain tasks, where the same views may not be guaranteed between vehicle designs. 
        
        We conclude that the late fusion technique is a strong baseline toward problems where multiple data streams, possibly under noise and dropped instances, are sampled simultaneously for continuous task inference.

\section*{Acknowledgment}

The authors would like to acknowledge LISA colleagues and research sponsors, especially Toyota CSRC, Qualcomm, and the valuable comments from distinguished colleagues from NHTSA.

\clearpage
{\small
\bibliographystyle{elsarticle-num.bst}
\bibliography{refs}

\begin{thebibliography}{10}
\expandafter\ifx\csname url\endcsname\relax
  \def\url#1{\texttt{#1}}\fi
\expandafter\ifx\csname urlprefix\endcsname\relax\def\urlprefix{URL }\fi
\expandafter\ifx\csname href\endcsname\relax
  \def\href#1#2{#2} \def\path#1{#1}\fi

\bibitem{zangi2022driver}
N.~Zangi, R.~Srour-Zreik, D.~Ridel, H.~Chasidim, A.~Borowsky, Driver
  distraction and its effects on partially automated driving performance: A
  driving simulator study among young-experienced drivers, Accident Analysis \&
  Prevention 166 (2022) 106565.

\bibitem{cdc_driving}
D.~o.~H. United~States, C.~f. D.~C. Human~Services, Prevention, Distracted
  driving, Centers for Disease Control and Prevention (2022).

\bibitem{dingus2016driver}
T.~A. Dingus, F.~Guo, S.~Lee, J.~F. Antin, M.~Perez, M.~Buchanan-King,
  J.~Hankey, Driver crash risk factors and prevalence evaluation using
  naturalistic driving data, Proceedings of the National Academy of Sciences
  113~(10) (2016) 2636--2641.

\bibitem{naujoks2018partial}
F.~Naujoks, S.~H{\"o}fling, C.~Purucker, K.~Zeeb, From partial and high
  automation to manual driving: Relationship between non-driving related tasks,
  drowsiness and take-over performance, Accident Analysis \& Prevention 121
  (2018) 28--42.

\bibitem{rangesh2021autonomous}
A.~Rangesh, N.~Deo, R.~Greer, P.~Gunaratne, M.~M. Trivedi, Autonomous vehicles
  that alert humans to take-over controls: Modeling with real-world data, in:
  2021 IEEE International Intelligent Transportation Systems Conference (ITSC),
  IEEE, 2021, pp. 231--236.

\bibitem{liang2022foundations}
P.~P. Liang, A.~Zadeh, L.-P. Morency, Foundations and recent trends in
  multimodal machine learning: Principles, challenges, and open questions,
  arXiv preprint arXiv:2209.03430 (2022).

\bibitem{seeland2021multi}
M.~Seeland, P.~M{\"a}der, Multi-view classification with convolutional neural
  networks, Plos one 16~(1) (2021) e0245230.

\bibitem{wang2021advances}
S.~Wang, M.~E. Celebi, Y.-D. Zhang, X.~Yu, S.~Lu, X.~Yao, Q.~Zhou, M.-G.
  Miguel, Y.~Tian, J.~M. Gorriz, et~al., Advances in data preprocessing for
  biomedical data fusion: An overview of the methods, challenges, and
  prospects, Information Fusion 76 (2021) 376--421.

\bibitem{yeong2021sensor}
D.~J. Yeong, G.~Velasco-Hernandez, J.~Barry, J.~Walsh, Sensor and sensor fusion
  technology in autonomous vehicles: A review, Sensors 21~(6) (2021) 2140.

\bibitem{ouhami2021computer}
M.~Ouhami, A.~Hafiane, Y.~Es-Saady, M.~El~Hajji, R.~Canals, Computer vision,
  iot and data fusion for crop disease detection using machine learning: a
  survey and ongoing research, Remote Sensing 13~(13) (2021) 2486.

\bibitem{silva2021multi}
B.~Silva, F.~R. Barbosa-Anda, J.~Batista, Multi-view fine-grained vehicle
  classification with multi-loss learning, in: 2021 IEEE international
  conference on autonomous robot systems and competitions (ICARSC), IEEE, 2021,
  pp. 209--214.

\bibitem{negrete2023multi}
S.~B. Negrete, H.~Arai, K.~Natsume, T.~Shibata, Multi-view image-based behavior
  classification of wet-dog shake in kainate rat model, Frontiers in Behavioral
  Neuroscience 17 (2023) 1148549.

\bibitem{khajwal2023post}
A.~B. Khajwal, C.-S. Cheng, A.~Noshadravan, Post-disaster damage classification
  based on deep multi-view image fusion, Computer-Aided Civil and
  Infrastructure Engineering 38~(4) (2023) 528--544.

\bibitem{wu2022ensemble}
L.~Wu, A.~Chen, P.~Salama, K.~W. Dunn, E.~J. Delp, An ensemble learning and
  slice fusion strategy for three-dimensional nuclei instance segmentation, in:
  Proceedings of the IEEE/CVF Conference on Computer Vision and Pattern
  Recognition, 2022, pp. 1884--1894.

\bibitem{chen2017multi}
X.~Chen, H.~Ma, J.~Wan, B.~Li, T.~Xia, Multi-view 3d object detection network
  for autonomous driving, in: Proceedings of the IEEE conference on Computer
  Vision and Pattern Recognition, 2017, pp. 1907--1915.

\bibitem{larsson2016fractalnet}
G.~Larsson, M.~Maire, G.~Shakhnarovich, Fractalnet: Ultra-deep neural networks
  without residuals, arXiv preprint arXiv:1605.07648 (2016).

\bibitem{liang2018deep}
M.~Liang, B.~Yang, S.~Wang, R.~Urtasun, Deep continuous fusion for multi-sensor
  3d object detection, in: Proceedings of the European conference on computer
  vision (ECCV), 2018, pp. 641--656.

\bibitem{vora2020pointpainting}
S.~Vora, A.~H. Lang, B.~Helou, O.~Beijbom, Pointpainting: Sequential fusion for
  3d object detection, in: Proceedings of the IEEE/CVF conference on computer
  vision and pattern recognition, 2020, pp. 4604--4612.

\bibitem{zhou2020end}
Y.~Zhou, P.~Sun, Y.~Zhang, D.~Anguelov, J.~Gao, T.~Ouyang, J.~Guo, J.~Ngiam,
  V.~Vasudevan, End-to-end multi-view fusion for 3d object detection in lidar
  point clouds, in: Conference on Robot Learning, PMLR, 2020, pp. 923--932.

\bibitem{sagi2018ensemble}
O.~Sagi, L.~Rokach, Ensemble learning: A survey, Wiley Interdisciplinary
  Reviews: Data Mining and Knowledge Discovery 8~(4) (2018) e1249.

\bibitem{chan1995comparative}
P.~K. Chan, S.~J. Stolfo, A comparative evaluation of voting and meta-learning
  on partitioned data, in: Machine Learning Proceedings 1995, Elsevier, 1995,
  pp. 90--98.

\bibitem{chawla2004learning}
N.~V. Chawla, L.~O. Hall, K.~W. Bowyer, W.~P. Kegelmeyer, Learning ensembles
  from bites: A scalable and accurate approach, The Journal of Machine Learning
  Research 5 (2004) 421--451.

\bibitem{rokach2008genetic}
L.~Rokach, Genetic algorithm-based feature set partitioning for classification
  problems, Pattern Recognition 41~(5) (2008) 1676--1700.

\bibitem{ting2011feature}
K.~M. Ting, J.~R. Wells, S.~C. Tan, S.~W. Teng, G.~I. Webb, Feature-subspace
  aggregating: ensembles for stable and unstable learners, Machine Learning
  82~(3) (2011) 375--397.

\bibitem{wen2017ensemble}
G.~Wen, Z.~Hou, H.~Li, D.~Li, L.~Jiang, E.~Xun, Ensemble of deep neural
  networks with probability-based fusion for facial expression recognition,
  Cognitive Computation 9~(5) (2017) 597--610.

\bibitem{ayad2010voting}
H.~G. Ayad, M.~S. Kamel, On voting-based consensus of cluster ensembles,
  Pattern Recognition 43~(5) (2010) 1943--1953.

\bibitem{deng2014ensemble}
L.~Deng, J.~Platt, Ensemble deep learning for speech recognition, in: Proc.
  interspeech, 2014.

\bibitem{brown2005managing}
G.~Brown, J.~L. Wyatt, P.~Tino, Y.~Bengio, Managing diversity in regression
  ensembles., Journal of machine learning research 6~(9) (2005).

\bibitem{lin2012parameter}
S.-W. Lin, S.-C. Chen, Parameter determination and feature selection for c4. 5
  algorithm using scatter search approach, Soft Computing 16~(1) (2012) 63--75.

\bibitem{raftery2005using}
A.~E. Raftery, T.~Gneiting, F.~Balabdaoui, M.~Polakowski, Using bayesian model
  averaging to calibrate forecast ensembles, Monthly weather review 133~(5)
  (2005) 1155--1174.

\bibitem{huang2023uncertainty}
T.~Huang, V.~Merwade, Uncertainty analysis and quantification in flood
  insurance rate maps using bayesian model averaging and hierarchical bma,
  Journal of Hydrologic Engineering 28~(2) (2023) 04022038.

\bibitem{tian2023data}
M.~Tian, H.~Fan, Z.~Xiong, L.~Li, Data-driven ensemble model for probabilistic
  prediction of debris-flow volume using bayesian model averaging, Bulletin of
  Engineering Geology and the Environment 82~(1) (2023) 1--16.

\bibitem{fei2023bayesian}
S.~Fei, Z.~Chen, L.~Li, Y.~Ma, Y.~Xiao, Bayesian model averaging to improve the
  yield prediction in wheat breeding trials, Agricultural and Forest
  Meteorology 328 (2023) 109237.

\bibitem{monteith2011turning}
K.~Monteith, J.~L. Carroll, K.~Seppi, T.~Martinez, Turning bayesian model
  averaging into bayesian model combination, in: The 2011 International Joint
  Conference on Neural Networks, IEEE, 2011, pp. 2657--2663.

\bibitem{gimelfarb2018reinforcement}
M.~Gimelfarb, S.~Sanner, C.-G. Lee, Reinforcement learning with multiple
  experts: A bayesian model combination approach, Advances in neural
  information processing systems 31 (2018).

\bibitem{sankar2005bayesian}
A.~Sankar, Bayesian model combination (baycom) for improved recognition, in:
  Proceedings.(ICASSP'05). IEEE International Conference on Acoustics, Speech,
  and Signal Processing, 2005., Vol.~1, IEEE, 2005, pp. I--845.

\bibitem{kim2012bayesian}
H.-C. Kim, Z.~Ghahramani, Bayesian classifier combination, in: Artificial
  Intelligence and Statistics, PMLR, 2012, pp. 619--627.

\bibitem{littlestone1994weighted}
N.~Littlestone, M.~K. Warmuth, The weighted majority algorithm, Information and
  computation 108~(2) (1994) 212--261.

\bibitem{bayrak2022deep}
G.~Bayrak, A.~K{\"u}{\c{c}}{\"u}ker, A.~Y{\i}lmaz, Deep learning-based
  multi-model ensemble method for classification of pqds in a hydrogen
  energy-based microgrid using modified weighted majority algorithm,
  International Journal of Hydrogen Energy (2022).

\bibitem{blum1997empirical}
A.~Blum, Empirical support for winnow and weighted-majority algorithms: Results
  on a calendar scheduling domain, Machine Learning 26~(1) (1997) 5--23.

\bibitem{braouezec2010committee}
Y.~Braouezec, Committee, expert advice, and the weighted majority algorithm: An
  application to the pricing decision of a monopolist, Computational Economics
  35~(3) (2010) 245--267.

\bibitem{liang2022multiviz}
P.~P. Liang, Y.~Lyu, G.~Chhablani, N.~Jain, Z.~Deng, X.~Wang, L.-P. Morency,
  R.~Salakhutdinov, Multiviz: Towards visualizing and understanding multimodal
  models, in: The Eleventh International Conference on Learning
  Representations, 2022.

\bibitem{beltran2021deep}
L.~V.~B. Beltr{\'a}n, J.~C. Caicedo, N.~Journet, M.~Coustaty, F.~Lecellier,
  A.~Doucet, Deep multimodal learning for cross-modal retrieval: One model for
  all tasks, Pattern Recognition Letters 146 (2021) 38--45.

\bibitem{wan2022editorial}
S.~Wan, Z.~Gao, H.~Zhang, C.~Xiaojun, C.~Chen, A.~Tefas, Editorial paper for
  pattern recognition letters vsi on cross model understanding for visual
  question answering, Pattern Recognition Letters 160 (2022) 9--10.

\bibitem{xue2023dynamic}
Z.~Xue, R.~Marculescu, Dynamic multimodal fusion, in: Proceedings of the
  IEEE/CVF Conference on Computer Vision and Pattern Recognition, 2023, pp.
  2574--2583.

\bibitem{ohn2013driver}
E.~Ohn-Bar, S.~Martin, M.~M. Trivedi, Driver hand activity analysis in
  naturalistic driving studies: challenges, algorithms, and experimental
  studies, Journal of Electronic Imaging 22~(4) (2013) 041119--041119.

\bibitem{ohn2014head}
E.~Ohn-Bar, S.~Martin, A.~Tawari, M.~M. Trivedi, Head, eye, and hand patterns
  for driver activity recognition (2014) 660--665.

\bibitem{8373883}
G.~Borghi, E.~Frigieri, R.~Vezzani, R.~Cucchiara, Hands on the wheel: A dataset
  for driver hand detection and tracking, in: 2018 13th IEEE International
  Conference on Automatic Face \& Gesture Recognition (FG 2018), 2018, pp.
  564--570.
\newblock \href {https://doi.org/10.1109/FG.2018.00090}
  {\path{doi:10.1109/FG.2018.00090}}.

\bibitem{eraqi2019driver}
H.~M. Eraqi, Y.~Abouelnaga, M.~H. Saad, M.~N. Moustafa, Driver distraction
  identification with an ensemble of convolutional neural networks, Journal of
  Advanced Transportation 2019 (2019).

\bibitem{shahverdy2020driver}
M.~Shahverdy, M.~Fathy, R.~Berangi, M.~Sabokrou, Driver behavior detection and
  classification using deep convolutional neural networks, Expert Systems with
  Applications 149 (2020) 113240.

\bibitem{weyers2019action}
P.~Weyers, D.~Schiebener, A.~Kummert, Action and object interaction recognition
  for driver activity classification (2019) 4336--4341.

\bibitem{yang2020refined}
L.~Yang, T.-Y. Yang, H.~Liu, X.~Shan, J.~Brighton, L.~Skrypchuk, A.~Mouzakitis,
  Y.~Zhao, A refined non-driving activity classification using a two-stream
  convolutional neural network, IEEE Sensors Journal 21~(14) (2020)
  15574--15583.

\bibitem{wang2021survey}
J.~Wang, W.~Chai, A.~Venkatachalapathy, K.~L. Tan, A.~Haghighat,
  S.~Velipasalar, Y.~Adu-Gyamfi, A.~Sharma, A survey on driver behavior
  analysis from in-vehicle cameras, IEEE Transactions on Intelligent
  Transportation Systems 23~(8) (2021) 10186--10209.

\bibitem{dang2019deep}
Q.~Dang, J.~Yin, B.~Wang, W.~Zheng, Deep learning based 2d human pose
  estimation: A survey, Tsinghua Science and Technology 24~(6) (2019) 663--676.

\bibitem{chen20222d}
H.~Chen, R.~Feng, S.~Wu, H.~Xu, F.~Zhou, Z.~Liu, 2d human pose estimation: A
  survey, arXiv preprint arXiv:2204.07370 (2022).

\bibitem{sun2019deep}
K.~Sun, B.~Xiao, D.~Liu, J.~Wang, Deep high-resolution representation learning
  for human pose estimation, in: Proceedings of the IEEE/CVF conference on
  computer vision and pattern recognition, 2019, pp. 5693--5703.

\bibitem{toshev2014deeppose}
A.~Toshev, C.~Szegedy, Deeppose: Human pose estimation via deep neural
  networks, in: Proceedings of the IEEE conference on computer vision and
  pattern recognition, 2014, pp. 1653--1660.

\bibitem{yang2017learning}
W.~Yang, S.~Li, W.~Ouyang, H.~Li, X.~Wang, Learning feature pyramids for human
  pose estimation, in: proceedings of the IEEE international conference on
  computer vision, 2017, pp. 1281--1290.

\bibitem{deo2019looking}
N.~Deo, M.~M. Trivedi, Looking at the driver/rider in autonomous vehicles to
  predict take-over readiness, IEEE Transactions on Intelligent Vehicles 5~(1)
  (2019) 41--52.

\bibitem{greer2023safe}
R.~Greer, N.~Deo, A.~Rangesh, P.~Gunaratne, M.~Trivedi, Safe control
  transitions: Machine vision based observable readiness index and data-driven
  takeover time prediction, arXiv preprint arXiv:2301.05805 (2023).

\bibitem{cummings2021safety}
M.~L. Cummings, B.~Bauchwitz, Safety implications of variability in autonomous
  driving assist alerting, IEEE Transactions on Intelligent Transportation
  Systems 23~(8) (2021) 12039--12049.

\bibitem{rangesh2018handynet}
A.~Rangesh, M.~M. Trivedi, Handynet: A one-stop solution to detect, segment,
  localize \& analyze driver hands (2018) 1103--1110.

\bibitem{yuen2019looking}
K.~Yuen, M.~M. Trivedi, Looking at hands in autonomous vehicles: A convnet
  approach using part affinity fields, IEEE Transactions on Intelligent
  Vehicles 5~(3) (2019) 361--371.

\bibitem{rangesh2021predicting}
A.~Rangesh, N.~Deo, R.~Greer, P.~Gunaratne, M.~M. Trivedi, Predicting take-over
  time for autonomous driving with real-world data: Robust data augmentation,
  models, and evaluation, arXiv preprint arXiv:2107.12932 (2021).

\bibitem{roitberg2022comparative}
A.~Roitberg, K.~Peng, Z.~Marinov, C.~Seibold, D.~Schneider, R.~Stiefelhagen, A
  comparative analysis of decision-level fusion for multimodal driver behaviour
  understanding, in: 2022 IEEE Intelligent Vehicles Symposium (IV), IEEE, 2022,
  pp. 1438--1444.

\bibitem{tawari2014robust}
A.~Tawari, M.~M. Trivedi, Robust and continuous estimation of driver gaze zone
  by dynamic analysis of multiple face videos, in: 2014 IEEE Intelligent
  Vehicles Symposium Proceedings, IEEE, 2014, pp. 344--349.

\bibitem{girshick2015fast}
R.~Girshick, Fast r-cnn, in: Proceedings of the IEEE international conference
  on computer vision, 2015, pp. 1440--1448.

\bibitem{lin2017feature}
T.-Y. Lin, P.~Doll{\'a}r, R.~Girshick, K.~He, B.~Hariharan, S.~Belongie,
  Feature pyramid networks for object detection, in: Proceedings of the IEEE
  conference on computer vision and pattern recognition, 2017, pp. 2117--2125.

\bibitem{He2015DeepRL}
K.~He, X.~Zhang, S.~Ren, J.~Sun, Deep residual learning for image recognition,
  2016 IEEE Conference on Computer Vision and Pattern Recognition (CVPR) (2015)
  770--778.

\bibitem{wang2019deep}
J.~Wang, K.~Sun, T.~Cheng, B.~Jiang, C.~Deng, Y.~Zhao, D.~Liu, Y.~Mu, M.~Tan,
  X.~Wang, W.~Liu, B.~Xiao, Deep high-resolution representation learning for
  visual recognition, TPAMI (2019).

\bibitem{greer2023multiview}
R.~Greer, L.~Rakla, A.~Gopalkrishnan, M.~Trivedi, Multi-view ensemble learning
  with missing data: Computational framework and evaluations using novel data
  from the safe autonomous driving domain (2023).
\newblock \href {http://arxiv.org/abs/2301.12592} {\path{arXiv:2301.12592}}.

\bibitem{sachdeva2023rank2tell}
E.~Sachdeva, N.~Agarwal, S.~Chundi, S.~Roelofs, J.~Li, B.~Dariush, C.~Choi,
  M.~Kochenderfer, Rank2tell: A multimodal driving dataset for joint importance
  ranking and reasoning, arXiv preprint arXiv:2309.06597 (2023).

\bibitem{dehghani2019subject}
A.~Dehghani, T.~Glatard, E.~Shihab, Subject cross validation in human activity
  recognition, arXiv preprint arXiv:1904.02666 (2019).

\bibitem{zhou2021domain}
K.~Zhou, Y.~Yang, Y.~Qiao, T.~Xiang, Domain adaptive ensemble learning, IEEE
  Transactions on Image Processing 30 (2021) 8008--8018.

\end{thebibliography}
}

\end{document}